\documentclass[journal,10pt]{IEEEtran}

\usepackage{multirow}
\usepackage{tabu}
\usepackage{makecell}
\usepackage{color}
\usepackage[table]{xcolor}
\usepackage{graphicx}
\usepackage{bm}
\usepackage{balance}
\usepackage{amsmath,amssymb,amsfonts}

\usepackage{hyperref}
\hypersetup{hidelinks=true}

\usepackage{amsmath,amssymb,amsfonts}
\usepackage{algorithmic}
\usepackage{graphicx}
\usepackage{textcomp}
\usepackage{balance}

\usepackage{multirow}
\usepackage{tabu}
\usepackage{makecell}
\usepackage[ruled,linesnumbered]{algorithm2e}
\SetAlCapNameFnt{\scriptsize}
\SetAlCapFnt{\scriptsize}

\definecolor{violet}{rgb}{0.56,0.0,0.56}

\def\BibTeX{{\rm B\kern-.05em{\sc i\kern-.025em b}\kern-.08em
    T\kern-.1667em\lower.7ex\hbox{E}\kern-.125emX}}
\begin{document}
\title{A Multi-Stage Automated Online Network Data Stream Analytics Framework for IIoT Systems}

\author{Li Yang, \IEEEmembership{Member, IEEE}, and Abdallah Shami, \IEEEmembership{Senior Member, IEEE}
\thanks{Li Yang and Abdallah Shami are with the Department of Electrical and Computer Engineering, Western University, London, ON N6A 3K7, Canada (e-mails: lyang339@uwo.ca; abdallah.shami@uwo.ca).}}

\maketitle

\begin{abstract}
Industry 5.0 aims at maximizing the collaboration between humans and machines. Machines are capable of automating repetitive jobs, while humans handle creative tasks. As a critical component of Industrial Internet of Things (IIoT) systems for service delivery, network data stream analytics often encounter concept drift issues due to dynamic IIoT environments, causing performance degradation and automation difficulties. In this paper, we propose a novel Multi-Stage Automated Network Analytics (MSANA) framework for concept drift adaptation in IIoT systems, consisting of dynamic data pre-processing, the proposed Drift-based Dynamic Feature Selection (DD-FS) method, dynamic model learning \& selection, and the proposed Window-based Performance Weighted Probability Averaging Ensemble (W-PWPAE) model. It is a complete automated data stream analytics framework that enables automatic, effective, and efficient data analytics for IIoT systems in Industry 5.0. Experimental results on two public IoT datasets demonstrate that the proposed framework outperforms state-of-the-art methods for IIoT data stream analytics. 
\end{abstract}


\markboth{Published in IEEE Transactions on Industrial Informatics}
{}

\IEEEpeerreviewmaketitle

\section{Introduction}
\label{sec:introduction}
\IEEEPARstart{T}{he} fourth industrial revolution (Industry 4.0) enables smart manufacturing via the application of various technologies, such as the Internet of Things (IoT), Artificial Intelligence (AI), big data analytics, cloud computing, and edge computing, robotics, and cybersecurity \cite{i5}. Industry 4.0 also achieves technological advancements that increase the level of automation in manufacturing facilities and warehouses \cite{i4}. Recently, as the fifth industrial revolution, Industry 5.0 has been proposed as a human-centered design solution for the next evolutionary state. In Industry 5.0, collaborative robots (cobots) and machines work collaboratively with human resources to enable customizable autonomous production through business social networks \cite{i5}. In Industry 5.0, humans can devote their creativity to responsible activities, while computers take over repetitive and monotonous duties, hence improving production quality and efficiency. Additionally, Industry 5.0 intends to increase the agility, efficiency, and scalability of production facilities and industries. It aims to enhance human-machine interaction via improved interfaces and automation systems programmed by human inventiveness, resulting in a multiple-fold increase in productivity \cite{i4}.

Network automation technologies are essential components in Industry 4.0 and 5.0, as well as 5G networks. Network automation refers to the process of automating the design, implementation, operation, and optimization of network services. Network automation can increase operational efficiency, reduce system errors, increase network service availability, and improve customer experience \cite{auto}. 

Network data analytics is a critical component of network automation systems. Automated data analytics driven by AI and Machine Learning (ML) algorithms provide insight into the present and future network activities. ML-driven network data analytics models can infer the purposes of network behaviors, conduct predictive analysis, and make decisions or recommendations. Thus, ML approaches provide promising solutions for network automation and 5G networks \cite{auto}. 

Industry 4.0 and Industry 5.0 both rely heavily on IoT systems. IoT is a network of machines, devices, sensors, and other technologies that connect to or interact with one another over the Internet \cite{iotyu}. The Industrial Internet of Things (IIoT) is a subcategory of IoT that refers to the deployment of IoT technology in industrial applications, such as manufacturing, transportation, healthcare, agriculture, etc. \cite{i4}. The IIoT's basic premise is that intelligent machines are often more effective and efficient than humans in properly capturing and analyzing data \cite{iiot}. 

On the other hand, IIoT data samples are usually non-stationary data streams generated in ever-changing IIoT environments due to their dynamic nature \cite{pwpae}. Thus, in real-world applications, IIoT data analytics often suffers from concept drift issues when IIoT data distributions change over time. The occurrence of concept drift poses considerable challenges in developing ML models, since their learning performance may progressively degrade owing to data distribution changes \cite{pwpae}. Thus, advanced online adaptive learning models should be developed to detect and react to concept drift that occurs in IIoT data streams. The drift adaptation procedure is also referred to as automated model updates in the network data analytics automation process, as its main purpose is to improve model performance by updating the learning model. 

In this work, a novel Multi-Stage Automated Network Analytics (MSANA) framework is proposed for IIoT data stream analytics and concept drift adaptation. It consists of four stages: dynamic data pre-processing, drift-based dynamic feature selection, base model learning and selection, and online ensemble model development. As a representative application of IIoT data analytics, the proposed framework is evaluated on two benchmark IoT anomaly detection datasets, IoTID20 \cite{iotid20} and CICIDS2017 datasets \cite{cicids2017}, to solve IIoT security problems.

The paper makes the following contributions:
\begin{enumerate}
\item It proposes MSANA, a novel and comprehensive framework for automated data stream analytics in IIoT systems, which includes typical data analytics procedures. The implementation code is open access on GitHub\footnote{
Code is available at: \url{https://github.com/Western-OC2-Lab/MSANA-Online-Data-Stream-Analytics-And-Concept-Drift-Adaptation}}.
\item It proposes the Window-based Performance Weighted Probability Averaging Ensemble (W-PWPAE) method, a novel ensemble drift adaptation strategy for online learning on dynamic data streams.
\item It proposes Drift-based Dynamic Feature Selection (DD-FS), a novel feature selection method, for data stream analytics with concept drift issues. 
\item It evaluates the proposed framework on two public IoT security datasets as a case study, and compares it with various state-of-the-art online learning approaches.
\end{enumerate}

To the best of our knowledge, no previous study has proposed such a complete pipeline/framework for automated data stream analytics in dynamic IIoT systems.

The remainder of the paper is organized as follows. Section II presents the related work about concept drift detection and adaptation. Section III describes the proposed multi-stage framework for automated data stream analytics. The experimental results are presented and discussed in Section IV. Section V discusses the potential deployment and the expected performance of the proposed framework in practical settings. Finally, Section VI summarizes the paper.

\section{Related Work}
Due to the dynamic nature of IIoT systems, network data analytics tasks often encounter concept drift issues when data distributions change over time, causing model learning performance degradation. To address concept drift, effective data stream analytics methods must be capable of detecting the occurrence of concept drift and then adapting to the detected concept drift. This section introduces and discusses existing methods for concept drift detection and adaptation.
\subsection{Concept Drift Detection}
There are two broad types of concept drift: sudden and gradual drifts \cite{drift1}. A sudden drift is a rapid change in the data distribution over a short period of time, while a gradual drift occurs when a new data distribution gradually replaces a historical concept. Different drift detection methods have been designed to detect different types of drift.

Distribution-based methods and performance-based methods are two common types of drift detection methods \cite{drift1}. ADaptive WINdowing (ADWIN) \cite{adwin} is a popular distribution-based approach that utilizes an adaptable sliding window to detect concept drift. ADWIN identifies data distribution changes by computing and comparing the characteristic values of the old and new distributions, such as the mean and variance values \cite{elena}. A significant change in the characteristic values over time indicates that a drift has occurred. Through the use of an adaptable sliding window, ADWIN works well with gradual drifts and long-term changes. However, changes in the statistics of data windows are sometimes virtual concept drift, resulting in false alarms and unnecessary model updates. 

Early Drift Detection Method (EDDM) \cite{eddm} is a widely-used performance-based drift detection method that tracks model performance changes based on the change rate of a learning model’s error rate and standard deviation by using a drift threshold and a warning threshold. If the error rate of a model increases dramatically, it will indicate model performance degradation and the occurrence of concept drift. Thus, EDDM can detect all real drifts that have degraded model performance, and is effective in detecting sudden drift. However, EDDM is inferior to distribution-based methods for gradual drift detection. 

\subsection{Concept Drift Adaptation}
After identifying a concept drift, learning models should be able to adapt to new concepts and enhance model performance. Existing drift-adaptive learning techniques fall into two primary categories: incremental learning and ensemble learning techniques. 

Incremental learning is the process of learning each incoming data sample in chronological order and partially updating the learner. Hoeffding Trees (HTs) \cite{drift1} is a basic incremental learning method that employs the Hoeffding inequality to determine the minimum number of data samples necessary for each split node, thus updating nodes to adapt to new samples. The Extremely Fast Decision Tree (EFDT) \cite{efdt} is a state-of-the-art incremental learning approach and an improved version of HTs. It selects and deploys each node split as soon as it reaches the confidence value, indicating a useful split. EFDT is able to adapt to concept drift more precisely and efficiently than HTs. Online Passive-Aggressive (OPA) \cite{opa} is another incremental learning algorithm that adapts to drift by passively reacting to correct predictions and aggressively responding to any errors. 

Several incremental methods are designed based on traditional ML algorithms. K-Nearest Neighbors with ADWIN drift detector (KNN-ADWIN) and Self-Adjusting Memory with KNN (SAM-KNN) \cite{samknn} are two improved versions of the traditional KNN model for online data analytics. KNN-ADWIN adds an ADWIN drift detector to the traditional KNN model and uses a dynamic window to determine which samples to retain for model updating. SAM-KNN, on the other hand, adapts to concept drift via the use of two memory modules: Short-Term Memory (STM) for the current concept and Long-Term Memory (LTM) for historical concepts \cite{samknn}. 

Ensemble online learning models are advanced drift-adaptive learning methods that integrate the output of multiple base learners for performance improvement. Leverage bagging (LB) \cite{lb} is a basic ensemble technique that constructs and combines multiple base learners (e.g., HTs) using bootstrap samples and the majority voting strategy. LB is simple to construct, but sensitive to noisy data. Adaptive Random Forest (ARF) \cite{arf} and Streaming Random Patches (SRP) \cite{srp} are two advanced ensemble online learning methods that train multiple HTs as base models and employs a drift detector (e.g., ADWIN) for each HT to address concept drift. ARF uses the local subspace randomization strategy to construct trees, while SRP uses global subspace randomization to generate random feature subsets for model learning. Using global subspace randomization improves the learning performance of SRP but increases its model complexity and learning time. 

Although ensemble online learning methods usually outperform incremental learning methods for dynamic data stream analytics, they are usually computationally expensive. Many deep learning methods are also unsuitable for data stream analytics due to their high complexity. Thus, despite the existence of many promising drift-adaptive learning methods, there is still much room for improvement.  The purpose of this study is to propose an ensemble framework capable of balancing model performance and learning speed. Additionally, existing approaches focus only on model learning but ignore other necessary data analytics procedures, such as data pre-processing and feature engineering. Thus, this paper proposes a comprehensive data analytics framework that includes other typical data analytics procedures. Moreover, we explored the automation of various data analytics procedures with the aim of network automation.

\section{Proposed Framework}

\subsection{System Overview}

\begin{figure}[!t]
\centerline{
\includegraphics[width=\columnwidth]{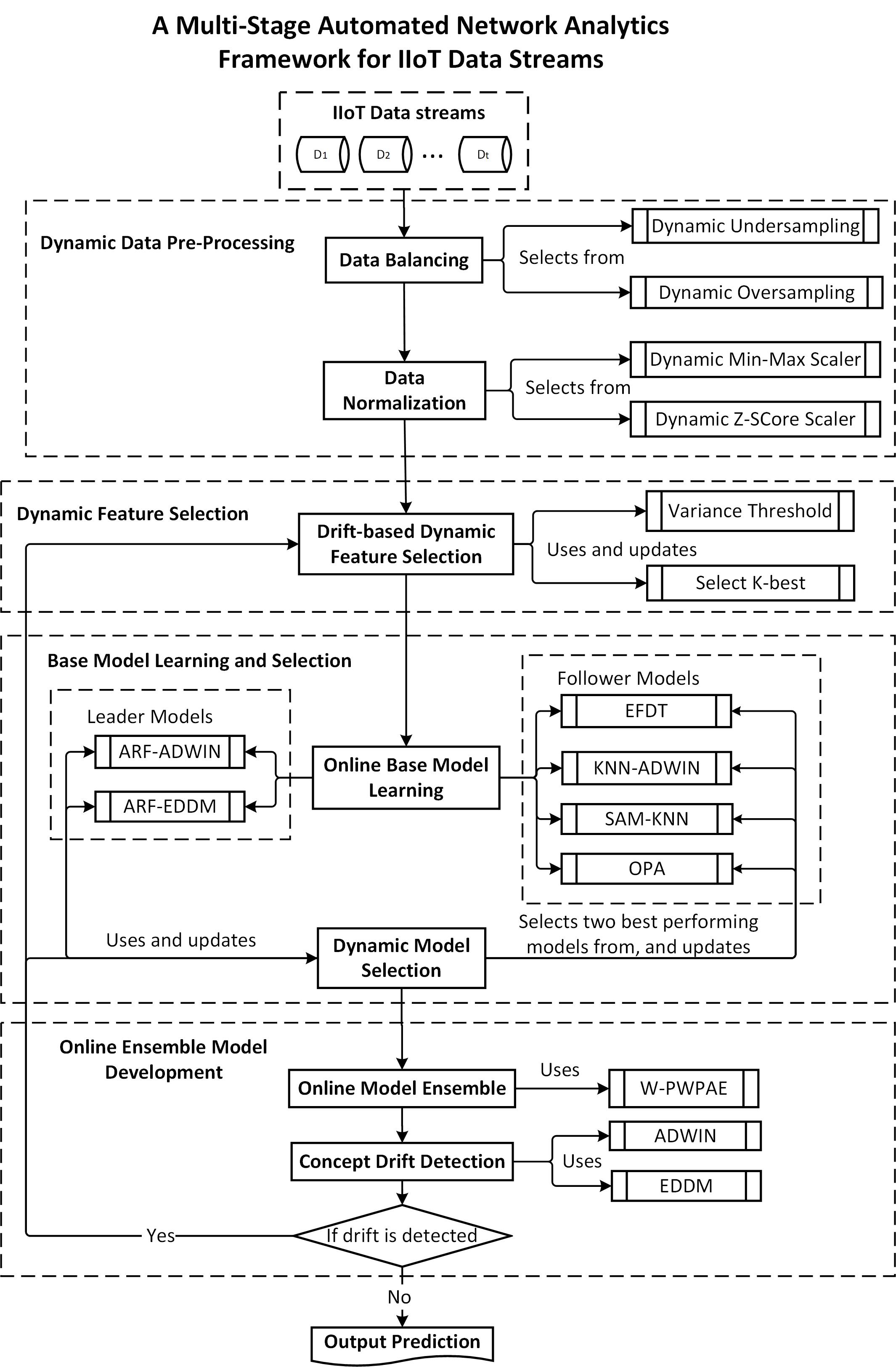}}
\caption{The framework of the proposed MSANA system.}
\label{fig1}
\end{figure}

The overview of the proposed MSANA framework for automated data analytics in IIoT systems is shown in Fig. \ref{fig1}. It consists of four primary stages: dynamic data pre-processing, drift-based dynamic feature selection, base model learning and selection, and online ensemble model development. 
At the first stage, to improve data quality, the incoming IIoT data streams are pre-processed using data balancing and normalization methods. The data balancing method is automatically selected from the dynamic undersampling and oversampling methods, depending on the specific performance and efficiency requirements. The details are discussed in Section III-B.1. The data normalization method is automatically selected from the dynamic min-max and z-score scalers, which are discussed in Section III-B.2. 
Secondly, dynamic feature selection methods are utilized to select the relevant features of data streams for specific tasks. The proposed Drift-based Dynamic Feature Selection (DD-FS) method consists of two basic feature selection methods: the variance threshold for discarding low-variance features and the select-k-best method for removing unimportant features. In DD-FS, the features are re-selected when a concept drift is detected, which is discussed in Section III-C.
The third phase consists of online base model learning and dynamic model selection. In online base model learning, six base online learners are trained using the cleaned data streams to perform basic data stream analytics. Specifically, ARF models with two different drift detectors, ARF-ADWIN and ARF-EDDM, are selected as the leader models, while the two  
best-performing models among four common online learning methods (i.e., EFDT, KNN-ADWIN, SAM-KNN, and OPA) are selected as the follower models. If concept drift occurs, the dynamic model selection module will re-select the base models according to new data distributions, and then update these learning models on the new concept samples in a time window. The details of base model learning and selection are discussed in Section III-D.1.
In the last phase, the proposed W-PWPAE method is used to integrate the output of the selected base learners based on their prediction probabilities and real-time error rates for the online model ensemble, which is discussed in Section III-D.2. The learning models are updated with the occurrence of concept drift, which is detected by two common drift detection methods, ADWIN and EDDM.
Finally, the prediction results of the ensemble learner on the processed data streams are delivered.

\subsection{Dynamic Data Pre-Processing}
Data pre-processing is mostly used to enhance the quality of data streams in order to increase model learning performance. In IIoT data streams, class imbalance and feature range difference are two potential data quality issues, which can be solved by data balancing and normalization methods, respectively. On the other hand, in dynamic IIoT systems, the class distributions and feature ranges are all dynamic variables that may change significantly over time. Thus, in contrast to the static process in traditional ML pipelines, data pre-processing in online data stream analytics should be dynamic procedures that must be updated on a continual basis.

\subsubsection{Data Balancing} \label{db}
Due to the dynamic nature of IIoT data streams, it is often difficult to maintain balanced distributions of all classes for classification problems, resulting in class imbalance issues. Training on imbalanced datasets may result in biased models with degraded performance. Resampling methods, including under-sampling and over-sampling methods, can be utilized to resolve class imbalance issues \cite{balancing}. The Dynamic Random Under-Sampling (DRUS) approach randomly discards data samples from majority classes to balance data. On the other hand, Dynamic Random Over-Sampling (DROS) balances data by continuously generating more samples for minority classes \cite{balancing}. 

The dynamic implementation of DRUS and DROS enables real-time updating of data distributions in response to class changes in dynamic data streams, thereby ensuring that the current data follows a balanced distribution. DRUS is fast and more suitable for IIoT systems that prioritize efficiency, while DROS is accurate and more suitable for IIoT systems that prioritize performance. As IIoT anomaly detection datasets are often highly imbalanced data with a tiny percentage of anomalies, the proposed system uses DROS to avoid discarding a significant proportion of majority class samples and omitting important information. When the ratio of the minority and the majority class samples reaches a given proportion threshold (e.g., 30\%), DROS is triggered to balance the data.

\subsubsection{Data Normalization}
ML and data analytics methods often prioritize features with higher values. Data scaling or normalization methods that normalize the features in a dataset to a comparable scale can often avoid biased models and improve learning performance. Z-score and min-max normalization are two commonly-used scaling techniques for data analytics problems \cite{norm}. 

The Z-score normalization method scales the feature value of each data sample $x$ to a normalized value $x_n$ \cite{norm}:
\begin{equation}
x_{n}=\frac{x-\tilde{\mu}}{\tilde{\sigma}}
\end{equation}
where $\tilde{\mu}$ and $\tilde{\sigma}$ are the real-time mean and standard deviation of all processed data samples. Unlike traditional normalization methods, $\tilde{\mu}$ and $\tilde{\sigma}$ are dynamic variables that are recalculated in real-time as new data samples arrive. When concept drift occurs, or the data distribution changes, its real-time mean and standard deviation can be changed accordingly.

In min-max normalization, the feature value of each data sample $x$ is scaled to \cite{norm}:
\begin{equation}
x_{n}=\frac{x-\widetilde{\min } }{\widetilde{\max }-\widetilde{\min }}
\end{equation}

Similar to Z-score normalization, $\widetilde{\min }$ and $\widetilde{\max }$ are the real-time minimum and maximum values of all processed data samples. Each time a new data sample is processed, $\widetilde{\min }$ and $\widetilde{\max }$ are updated based on the processed samples' values. Thus, normalized values for the processed samples can also be updated in response to data distribution changes. Min-max scaler can normalize all features to the same scale of 0-1.

Min-max normalization is better suitable for anomaly detection issues due to its ability to retain outliers (e.g., extremely large or small values) in datasets. On the other hand, Z-score normalization is robust to outliers, so it often performs well for other non-outlier-related data analytics problems. Thus, the normalization method can be determined according to specific problems. As the proposed system aims to solve IIoT anomaly detection problems as a use case, min-max normalization is selected for the proposed framework.

\subsection{Drift-based Dynamic Feature Selection}
As the original features are usually not the optimal features for specific tasks, the primary objective of feature selection is to return the updated data with optimal input features for performance improvement. Feature selection can also improve learning efficiency by discarding irrelevant and noisy features. The proposed Drift-based Dynamic Feature Selection (DD-FS) method used in this work consists of two feature selection techniques: variance threshold and select-k-best. 

Variance threshold is a feature selection method that aims to eliminate all low-variance features. The variance $\sigma^{2}$ of each feature can be denoted by \cite{river}:
\begin{equation}
\sigma^{2}=\frac{\sum_{i=1}^{n}\left(x_{i}-\bar{x}\right)^{2}}{n}
\end{equation}
where $n$ is the number of processed samples, $x$ is an input feature, and $\bar{x}$ denotes the mean value of $x$. 
Using the variance threshold approach can remove the features whose variance is lower than a given threshold. A low variance indicates that the corresponding feature is often uninformative as it has the same values across the majority of data samples. By removing low-variance features, the learning efficiency of models can be increased.

Select-k-best is a popular feature selection method in which the correlations between each input feature and the target variable are calculated as feature importance scores, and then the features with the $k$ highest importance scores are selected \cite{fs}. The feature importance scores can be computed using the Pearson correlation coefficient, a commonly-used metric to measure the correlations between two variables. It can be denoted by \cite{fs}:
\begin{equation}
Corr_{x y}=\frac{\sum_{i=1}^{n}\left(x_{i}-\bar{x}\right)\left(y_{i}-\bar{y}\right)}{\sqrt{\sum_{i=1}^{n}\left(x_{i}-\bar{x}\right)^{2}\left(y_{i}-\bar{y}\right)^{2}}}
\end{equation}
where $x$ is an input feature, $y$ is the target variable, $\bar{x}$ and $\bar{y}$ are the mean values of an input feature and the target variable, respectively. 

The Pearson correlation coefficient has a range of -1.0 to 1.0, with -1.0 indicating a perfect negative relationship, 1.0 indicating a perfect positive relationship, and 0 indicating that the two variables are fully uncorrelated. Thus, it can be used to quantify the strength of a relationship between each feature and the target variable, making it easy to assess and compare the importance of different features.

Algorithm 1 illustrates the primary procedures of DD-FS. At the first stage, the initial training set $S_{train}$ is learned by the variance threshold (${MF}_{1}$) and the select-k-best method (${MF}_{2}$) to generate the initial optimal feature set $F^\prime$. The initial $F^\prime$ is then utilized to update the feature list of each incoming data sample $x_i$ from the online test set $S_{test}$. If a drift is detected by the drift detectors, the variance threshold and select-k-best feature selection methods will re-learn the recent data samples in a time window $s$ as the new concept samples to generate an updated optimal feature set $F^\prime$. During the entire online data stream analytics process, this drift-based feature re-selection procedure is automatically repeated each time a concept drift is detected. The drift adaptation functionality of DD-FS is based on the assumption that when concept drift occurs and data distribution changes, the best suitable feature set will change as well.
\begin{algorithm}[t!]
    {\scriptsize
	\caption{Drift-based Dynamic Feature Selection (DD-FS)}
	\label{algo:event}
	\LinesNumbered
	\KwIn{
	\\\quad $F$: the original feature set.}
	\KwOut{
	\\\quad $F^\prime$: the updated feature set.}
	${MF}_{1},F^\prime \leftarrow$ VarianceThreshold\_Learning$(S_{train},F)$;\\
    $MF_{2},F^\prime \leftarrow$ SelectKBest\_Learning$(S_{train},F^\prime)$; \tcp*[f]{Apply two FS methods on the training set}\\ 
    $S_{train}^{\prime} \leftarrow S_{train}.$Update$(F^\prime)$; \tcp*[f]{Update the feature list}\\
	$Drift = 0$; \tcp*[f]{Drift indicator}\\
	\For{\rm{each data instance} $x_i\in S_{test}$}{	
    	$x_{i}^{\prime} \leftarrow MF_{1}.$Transform$(x_{i},F^\prime)$;\\
    	$x_{i}^{\prime} \leftarrow MF_{1}.$Transform$(x_{i}^{\prime},F^\prime)$;\\
        $Drift_{1} \leftarrow $ADWIN.Update$()$;\\
        $Drift_{2} \leftarrow $EDDM.Update$()$;\\
        
	    \If(\tcp*[f]{If both detectors have detected drifts }){$(Drift_{1}==1)\&\&(Drift_{2}==1)$ } 
    	    {
    	    $S_{new} \leftarrow S_{test}[i-s, i]$; \tcp*[f]{Recent window samples}\\
    	    ${MF}_{1},F^\prime \leftarrow$ VarianceThreshold\_Learning$(S_{new},F^\prime)$;\\
            $MF_{2},F^\prime \leftarrow$ SelectKBest\_Learning$(S_{new},F^\prime)$;\\
    	    $S_{new}^{\prime} \leftarrow S_{new}$.Update$(F^\prime)$; \tcp*[f]{Re-select features}\\
    	    }
    }
    \KwRet $F^\prime$; \tcp*[f]{Return the updated feature set}\\

}
\end{algorithm}

\begin{algorithm}[t!]
    {\scriptsize
	\caption{Base Model Learning and Dynamic Selection}
	\label{algo2}
	\LinesNumbered
	\KwIn{ 
    $M = \{M_1,M_2,M_3,M_4,M_5,M_6\}$: the initial base model set, $M_1$ = ARF-ADWIN, $M_2$ = ARF-EDDM, $M_3$ = EFDT, $M_4$ = KNN-ADWIN, $M_5$ = SAM-KNN, $M_6$ = OPA;
    }
	\KwOut{
	$M^{\prime} = \{M_{l1},M_{l2},M_{f1},M_{f2}\}$: the selected and trained base models;
	}
	$Drift = 0$; \tcp*[f]{Drift indicator}\\
    $M \leftarrow $Models\_LearningBatch$(M,S_{train}^{\prime})$; \tcp*[f]{Train six initial base learners on the training set}\\
    $M^{\prime} = \{M_{l1},M_{l2},M_{f1},M_{f2}\} \leftarrow $ Model\_Selection$(M)$; \tcp*[f]{Select the base learners based on their performance, $M_{l1}$ and $M_{l2}$ are the leader models (ARF-ADWIN and ARF-EDDM), $M_{f1}$ and $M_{f2}$ are the two best performing models among the other four base models}\\
    
	\For{\rm{each data instance} $x_i\in S_{test}$}{	
        $y_{pred_i} \leftarrow $ Models\_Prediction$(M^{\prime},x_i)$; \tcp*[f]{Use the four selected base learners to predict each new sample}\\
        $M^{\prime} \leftarrow $Model\_LearningOne$(M^{\prime},x_i)$; \tcp*[f]{The four selected base learners learn each new sample}\\

        $Drift_{1} \leftarrow $ADWIN.Update$(y_{pred_i}, y_{true_i})$;\\
        $Drift_{2} \leftarrow $EDDM.Update$(y_{pred_i}, y_{true_i})$;\\
        
	    \eIf(\tcp*[f]{If both detectors have detected drifts }){$(Drift_{1}==1)\&\&(Drift_{2}==1)$ } 
    	    {
            $M^{\prime} = \{M_{l1},M_{l2},M_{f1},M_{f2}\} \leftarrow $ Model\_Selection$(M^{\prime})$; \tcp*[f]{Re-select the base learners based on their real-time window performance}\\
            $S_{new} \leftarrow S_{test}[i-s, i]$; \tcp*[f]{Recent window samples}\\
            $M^{\prime} \leftarrow $Model\_LearningBatch$(M^{\prime},S_{new})$; \tcp*[f]{Update the learners on new concept data samples}\\
    	    }
	    {
            $M^{\prime} \leftarrow $Model\_LearningOne$(M^{\prime},x_i)$; \tcp*[f]{The four selected base learners learn the new sample}\\
	    }
    }
    \KwRet $M^\prime$; \tcp*[f]{Return the selected model set}\\

}
\end{algorithm}

\subsection{Model Learning}
The proposed model learning framework consists of two stages: base model learning \& selection and online model ensemble. The base model learned in the first stage is selected to construct an ensemble model in the second stage.
\subsubsection{Base Model Learning and Dynamic Selection}
For base model learning, the lightweight online learning methods introduced in Section II-B are used to learn the data streams. At the initial stage, the learning methods process a small-size training set to generate initial base learners. The learners then learn and predict each incoming sample from the online test set and update themselves if concept drift occurs. 

Appropriate base learners should be selected in the proposed dynamic model selection process. Model selection is the process of selecting appropriate base models to construct a robust ensemble model. The specifics of the base model learning and dynamic selection process in the proposed framework are shown in Algorithm 2. As the proposed framework is designed for real-time IIoT systems, it should strike a balance between learning performance and efficiency. Thus, relatively lightweight models are selected from the initial base model set $M$ as the base models in the proposed framework. Two leader models and two follower models are selected as the model set $M^{\prime}$ in the proposed system.

Two drift detectors introduced in Section II-A, ADWIN and EDDM, are used together to detect concept drift in the proposed framework. ADWIN identified concept drift by comparing the mean values of the two consecutive data windows, and EDDM detects drift by monitoring the model performance change through a drift threshold $\beta$ and a warning threshold $\alpha$, denoted by \cite{eddm}:
\begin{equation}
\left\{\begin{array}{l}
\text{if} (p_{t}^{\prime}+2 * s_{t}^{\prime}) / (p_{\max}^{\prime}+2 * s_{\max}^{\prime}) < \alpha \rightarrow \text {Warning;} \\
\text{if} (p_{t}^{\prime}+2 * s_{t}^{\prime}) / (p_{\max}^{\prime}+2 * s_{\max}^{\prime}) < \beta \rightarrow \text {Drift,}
\end{array}\right.
\end{equation}
where $p_{t}^{\prime}$ and $p_{\max}$ are the running average and maximum error rate changes at time $t$, while $(s_{t}^{\prime}$ and $s_{\max}$ are the running average and maximum standard deviations. As ADWIN works well with gradual drift and EDDM is effective in detecting sudden drift, using both of them enables the detection of different types of concept drift. 

As introduced in Section II-B, Hoeffding Tree (HT) is a basic online learner that adapts to concept drift using the Hoeffding bound $\epsilon$, defined as \cite{efdt}: 
\begin{equation}\epsilon=\sqrt{\frac{R^2 \ln (1 / \delta)}{2 n}},\end{equation}
where the split on the best feature of range $R$ after processing $n$ data samples is the same as if the model had observed an infinite number of samples with probability $1-\delta$.

ARF \cite{arf} is an advanced ensemble model that uses local subspace randomization to construct HTs for drift adaptation and employs a drift detector for drift detection. ARF has been proven to be both efficient and effective for a variety of data stream analytics problems \cite{pwpae}. Thus, ARF with two different drift detectors, ARF-ADWIN and ARF-EDDM, are selected as the two leader base models. Moreover, due to the high effectiveness of ARF, using two ARF models with different drift detectors as leader models enables the proposed ensemble model to retain high accuracy even when the follower models do not perform as well as the leader models.  

Next, the two follower models are chosen from the other four lightweight and state-of-the-art online learning methods introduced in Section II-B: EFDT \cite{efdt}, KNN-ADWIN \cite{samknn}, SAM-KNN \cite{samknn}, and OPA \cite{opa}, as shown in Fig. \ref{fig1}. Although these four models are not as effective as ARF, they are all well-performing and fast adaptive online learning models that can address concept drift efficiently. The other three models introduced in Section II-B, LB \cite{lb}, SRP \cite{srp}, and PWPAE \cite{pwpae}, are not selected in the proposed ensemble framework due to their high computational complexity. They are used as comparison models in the experiments. 

After selecting the two leader models and two follower models in $M^{\prime}$, they are used to predict and learn each incoming data sample, as shown in Algorithm 2.
If concept drift occurs, the follower models will be re-selected from the four candidate models based on their real-time performance in the sliding window of data. This process is called dynamic model selection. The selected two best-performing follower models on the data of the new concept are then combined with the ARF-ADWIN and ARF-EDDM leader models to construct a new ensemble model. Additionally, after detecting a concept drift using ADWIN and EDDM, all four base models learn the most recent sliding window of data to construct the updated base models that can adapt to the new concept. This dynamic model selection and learning process are beneficial to concept drift adaptation because it enables more effective model updates when the data distribution changes.

\subsubsection{Online Ensemble Model Development}
After obtaining the base learning models, their prediction outputs are then integrated to construct an ensemble model with improved performance using a novel ensemble strategy, named Window-based Performance Weighted Probability Averaging Ensemble (W-PWPAE). It is extended from the ensemble method PWPAE proposed by the authors in \cite{pwpae}. 

W-PWPAE integrates base models by assigning dynamic weights to the prediction probabilities of the base models, and then averaging the weighted probabilities. The class with the highest mean probability value, indicating the most confident result, is then selected as the final prediction result. Assuming a data stream $D=\{(x_1,y_1 ),\dots,(x_n,y_n )\}$, and there are $c$ different classes for the target variable, $y \in {1,\dots,c}$, the predicted target class for each input data $x$ can be denoted by:
\begin{equation}
\hat{y}=\underset{i \in\{1, \cdots, c\}}{\operatorname{argmax}} \frac{\sum_{j=1}^{b} w_{j} p_{j}\left(y_{j}=i \mid L_{j}, x_{j}\right)}{b}
\end{equation}
where $L_j$ represents the $j_{th}$ base model, $p_j (y_{j}=i|L_j,x_{j})$ indicates the prediction probability of a class value $i$ on the data sample $x$ using the $j_{th}$ base learner $L_j$; $b$ is the number of base models, where $b=4$ for the proposed ensemble model, and $w_j$ represents the weight of each base model $L_j$.

The weight $w_j$ of each base model $L_j$ is computed based on the reciprocal of the model’s real-time error rate in the latest window $s$:
\begin{equation}w_{j}=\frac{1}{ Error_{s, j}+\epsilon}\end{equation}
where $\epsilon$ is a small constant used to avoid a denominator of 0. Thus, higher weights are assigned to the base models with lower error rates and better performance.

The window error rate for each base model $L_j$ can be calculated by: 
\begin{equation}
{ Error }_{s, j}=\frac{1}{s} \sum_{k=1}^{s} \delta\left(L_{j}\left(x_{k}\right), y_{k}\right),
\end{equation}
where $s$ is the sliding window size, and $\delta\left(L_{j}\left(x_{k}\right), y_{k}\right)$ is the loss function calculated based on the predicted value $L_j (x_k )$ and the ground-truth value $y_k$. 

The window size $s$ is determined according to the detected drift information:
\begin{equation}
\left\{\begin{array}{l}
 s=DriftArr[-1] \text {, if }DriftArr\text { is not empty; } \\
 s=\alpha * N \text {, if }DriftArr\text { is empty, }
\end{array}\right.
\end{equation}
where $DriftArr$ is an array used to record the index of all drift points, $\alpha$ is a ratio, and $N$ represents the total number of processed samples. If any concept drift is detected, the window will include all the data samples with an index from the last drift point $DriftArr[-1]$ to the current data point, indicating the new concept data. Otherwise, the window size is determined by a proportion of processed data samples, $\alpha * N$. For example, if $\alpha$ is set to 0.1, the window will contain the most recent 10\% of the processed data samples. The computational complexity of the W-PWPAE ensemble model is primarily determined by the complexity of the selected based models, whereas the W-PWPAE method itself has a low computational complexity of $O(sck)$, where the window size $s$, the number of distinct classes $c$, and the number of base learners $k$, all usually have small values.

In comparison to other existing drift-adaptive online learning methods, the proposed W-PWPAE approach has the following advantages: 
\begin{enumerate}
\item Unlike many other existing ensemble learning methods that use the hard majority voting strategy, the proposed framework uses the confidence probability of each base classifier for each class, a more robust and flexible strategy. It considers each base classifier’s uncertainty for each data sample to prevent arbitrary decisions. 
\item Using the window and performance-based dynamic weighting strategy enables the proposed framework to focus on the model performance on the new concept data, leading to a more effective concept drift adaptation. 
\item The selection of lightweight base learning models enables the construction of an efficient ensemble model, as the primary drawback of many existing ensemble models is their high complexity.
\end{enumerate}

\section{Performance Evaluation}

\subsection{Experimental Setup}
The proposed system was implemented in Python 3.7 by extending the River \cite{river} library on a computer equipped with an i7-8700 CPU and 16 GB of memory, representing an IIoT cloud server machine for large data stream analytics. 

\begin{figure}[!t]
\centerline{
\includegraphics[width=\columnwidth]{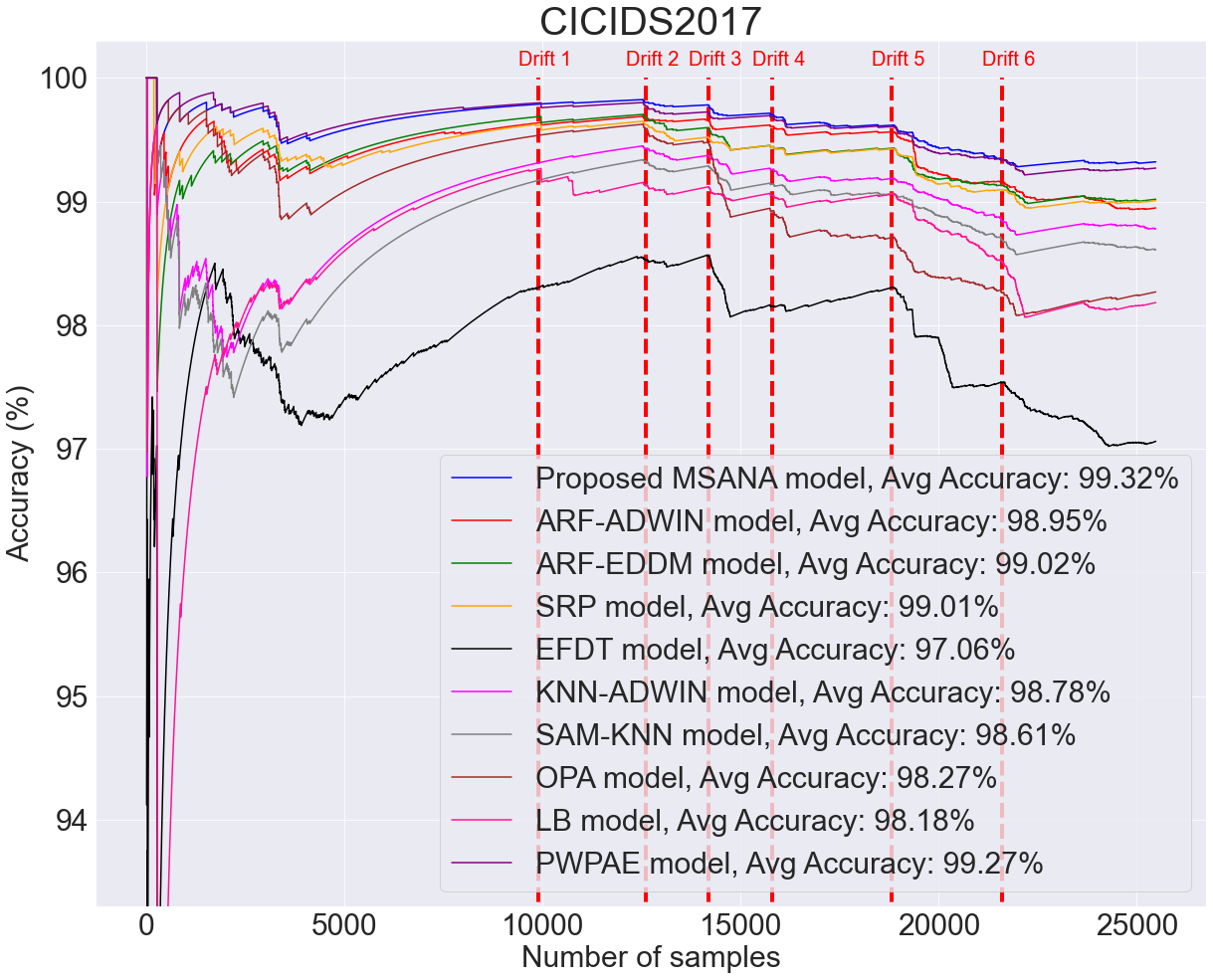}}
\caption{Accuracy comparison of state-of-the-art drift adaptation methods on the CICIDS2017 dataset.}
\label{cic}
\end{figure}

The proposed approach is evaluated on two public IIoT security datasets: IoTID20 \cite{iotid20} and CICIDS2017 \cite{cicids2017}. IoTID20 is a relatively new IoT dataset that was created by generating IoT network traffic data from both legitimate and malicious IoT devices, including 83 different network features. CICIDS2017 is a public network security dataset that was contributed by the Canadian Institute of Cybersecurity and contained state-of-the-art cyberattack scenarios. Due to the fact that the CICIDS2017 dataset was created by launching a variety of different attack types in different time periods, the attack patterns in the dataset have changed over time, resulting in six concept drifts, as shown in Figure \ref{cic}. For the purpose of this work, A representative IoTID20 subset with 6,252 data and a sampled CICIDS2017 subset with 28,303 records are utilized for the model evaluation. 

The IIoT data analytics use case solved by the proposed system is anomaly detection, which can be regarded as a binary classification problem by labeling each data sample in the two datasets as a normal sample or an attack sample. The proposed framework is evaluated using the combination of hold-out and prequential validations. For hold-out validation, the first 10\% of data is utilized for training the initial base models, and the remaining 90\% is used for online testing of dynamic data streams. Prequential validation, also known as test-and-train validation, is utilized to evaluate the proposed model for online learning. In prequential validation, each input sample in the online test set is firstly tested by the learners to monitor their real-time performance, and then learned by the learning model for potential model updates \cite{pwpae}.

To provide a comprehensive analysis of the experimental results, the proposed framework is evaluated from multiple perspectives. Firstly, from the effectiveness perspective, four performance metrics, including accuracy, precision, recall, and F1-score, are used to evaluate the proposed framework. As the IoT anomaly detection datasets are imbalanced, precision, recall, and F1-score are used with accuracy to provide a comprehensive view of the model performance. Secondly, from the efficiency perspective, two ML and data analytics-related Quality of Service (QoS) parameters, latency and throughput \cite{throughput}, are used to assess the proposed framework's learning efficiency. Latency indicates the inference time of learning models, and is computed by the average inference/test time per packet/data sample. Throughput refers to the number of processed data samples/packets in one unit of time (e.g., second). Low latency and high throughput are two primary performance requirements for ML and data analytics models \cite{throughput}. Moreover, the Probability Density Function and contributors (model components) of the latency are also used to evaluate the efficiency of the proposed model. To achieve real-time analytics, efficient learning models should strike a balance between prediction accuracy and latency.

\begin{figure}[!t]
\centerline{
\includegraphics[width=\columnwidth]{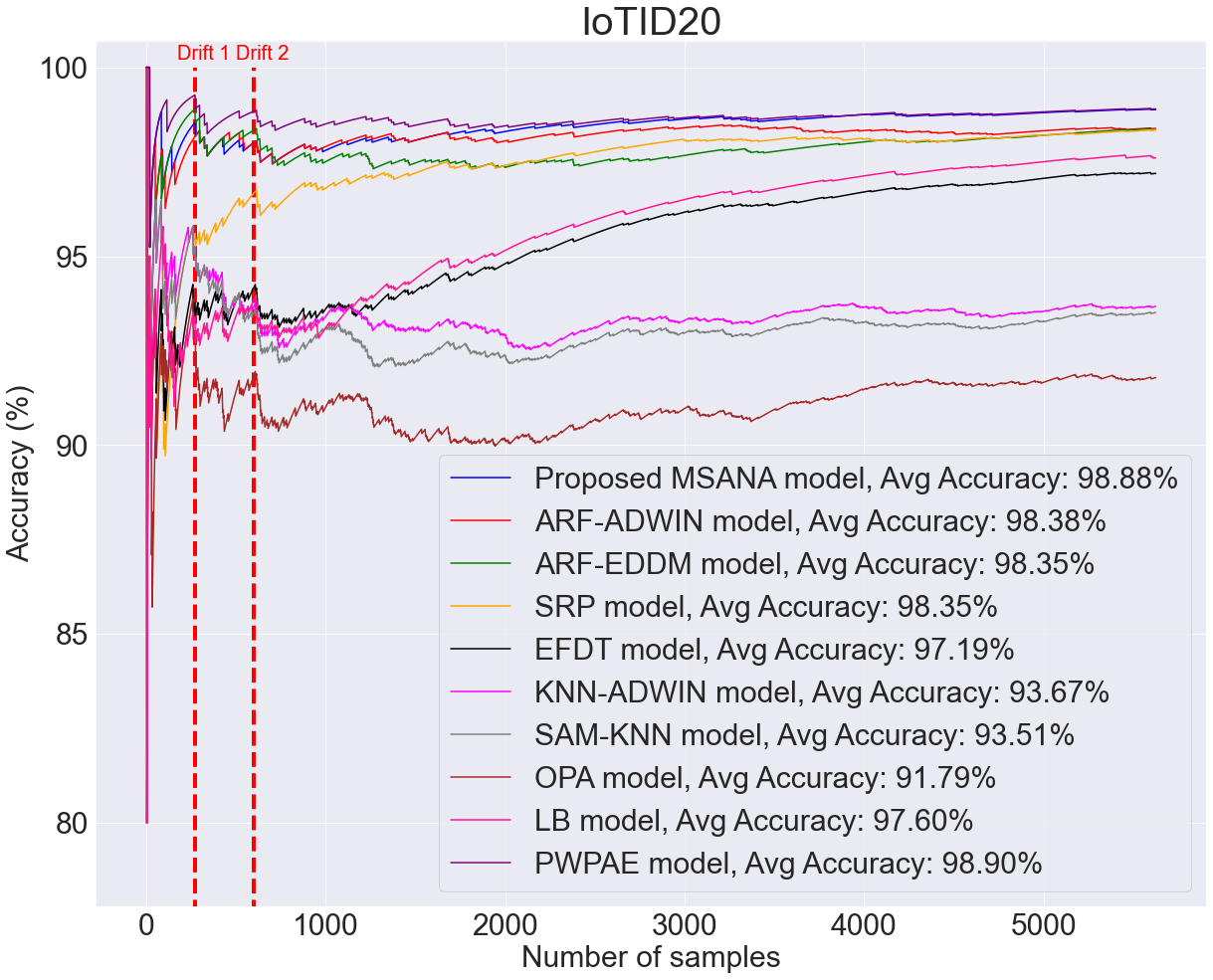}}
\caption{Accuracy comparison of state-of-the-art drift adaptation methods on the IoTID20 dataset.}
\label{iotid}
\end{figure}

\begin{table*}[]
\centering%
\caption{Performance Comparison of State-of-The-Art Drift Adaptive Online Learning Methods}
\setlength\extrarowheight{1pt}
\scalebox{0.80}{
\begin{tabular}{|>{\centering\arraybackslash}m{9em}|>{\centering\arraybackslash}m{4em}|>{\centering\arraybackslash}m{4em}|>{\centering\arraybackslash}m{3.5em}|>{\centering\arraybackslash}m{3em}|>{\centering\arraybackslash}m{4.5em}|>{\centering\arraybackslash}m{5.8em}|>{\centering\arraybackslash}m{4em}|>{\centering\arraybackslash}m{4em}|>{\centering\arraybackslash}m{3.5em}|>{\centering\arraybackslash}m{3em}|>{\centering\arraybackslash}m{4.5em}|>{\centering\arraybackslash}m{5.8em}|}
\hline
\multirow{4}{*}{\textbf{Method}} & \multicolumn{6}{c|}{\textbf{CICIDS2017 Dataset}}                                & \multicolumn{6}{c|}{\textbf{IoTID20 Dataset}}                                        \\ \cline{2-13} 
                                 & \textbf{Accuracy (\%)} & \textbf{Precision (\%)} & \textbf{Recall (\%)} & \textbf{F1 (\%)} & \textbf{Latency/ Inference Time (ms)} & \textbf{Throughput (samples per second)} & \textbf{Accuracy (\%)} & \textbf{Precision (\%)} & \textbf{Recall (\%)} & \textbf{F1 (\%)} & \textbf{Latency/ Inference Time (ms)} & \textbf{Throughput (samples per second)}  \\ 
\hline
ARF-ADWIN \cite{arf}                        & 98.95                  & 97.50                   & 96.12                & 96.81            & 1.12                                  & 892.86                                    & 98.38                  & 98.55                   & 99.75                & 99.15            & 1.09                                  & 917.43                                     \\ 
\hline
ARF-EDDM \cite{arf}                         & 99.02                  & 97.72                   & 96.33                & 97.02            & 1.03                                  & 970.87                                    & 98.35                  & 98.44                   & 99.83                & 99.13            & 0.85                                  & 1176.47                                    \\ 
\hline
SRP \cite{srp}                              & 99.01                  & 97.63                   & 96.36                & 96.99            & 3.75                                  & 266.67                                    & 98.35                  & 98.55                   & 99.72                & 99.13            & 3.66                                  & 273.22                                     \\ 
\hline
EFDT \cite{efdt}                              & 97.06                  & 91.31                   & 90.94                & 91.13            & 0.48                                  & 2083.33                                   & 97.19                  & 97.53                   & 99.55                & 98.53            & 0.43                                  & 2325.58                                    \\ 
\hline
KNN-ADWIN \cite{samknn}                        & 98.78                  & 95.52                   & 97.27                & 96.36            & 0.71                                  & 1408.45                                   & 93.67                  & 95.30                   & 98.13                & 96.69            & 0.54                                  & 1851.85                                    \\ 
\hline
SAM-KNN \cite{samknn}                          & 98.61                  & 94.88                   & 96.86                & 95.86            & 1.09                                  & 917.43                                    & 93.51                  & 94.93                   & 98.38                & 96.62            & 1.38                                  & 724.64                                     \\ 
\hline
OPA \cite{opa}                              & 98.27                  & 94.80                   & 94.77                & 94.79            & 0.32                                  & 3125.0                                    & 91.79                  & 95.99                   & 95.27                & 95.63            & 0.21                                  & 4761.9                                     \\ 
\hline
LB \cite{lb}                               & 98.18                  & 96.28                   & 96.07                & 96.18            & 4.26                                  & 234.74                                    & 97.60                  & 98.28                   & 99.19                & 98.73            & 4.58                                  & 218.34                                     \\ 
\hline
PWPAE \cite{pwpae}                            & 99.27                  & 98.40                   & 97.19                & 97.79            & 7.44                                  & 134.41                                    & 98.90                  & 98.86                   & 99.98                & 99.42            & 9.08                                  & 110.13                                     \\ 
\hline
\textbf{Proposed MSANA}                   & \textbf{99.32}                  & \textbf{98.31}                   & \textbf{97.59}                & \textbf{97.95}            & \textbf{3.53}                                  & \textbf{283.29}                                    & \textbf{98.88}                  & \textbf{98.88}                   & \textbf{99.94}                & \textbf{99.41}            & \textbf{2.81}                                  & \textbf{355.87}                                     \\
\hline

\end{tabular}
}
\label{results}
\end{table*}

\subsection{Experimental Results and Discussion}
Figures \ref{cic} \& \ref{iotid} and Table \ref{results} illustrate the performance comparison of the proposed MSANA method against other state-of-the-art online adaptive learning methods presented in Section II-B, including ARF-ADWIN \cite{arf}, ARF-EDDM \cite{arf}, SRP \cite{srp}, EFDT \cite{efdt}, KNN-ADWIN \cite{samknn}, SAM-KNN \cite{samknn}, OPA \cite{opa}, LB \cite{lb}, and PWPAE \cite{pwpae}.

As shown in Fig. \ref{cic} and Table \ref{results}, on the CICIDS2017 dataset, the two leader models, ARF-ADWIN and ARF-EDDM, achieve high accuracy of 98.95\% and 99.02\%. This demonstrates why they were chosen as leader learners. Among the four follower models (EFDT, KNN-ADWIN, SAM-KNN, and OPA), SAM-KNN and KNN-ADWIN are the two best-performing models on the CICIDS2017 dataset, so they are selected as the two final follower models to construct the initial ensemble model. After using the two leader models and two follower models to build an ensemble learner, the proposed MSANA achieves the highest accuracy of 99.32\% and F1-score of 97.95\% among all the evaluated online learning models. Additionally, the inference time of MSANA is only 3.53 ms per packet, which is less than for other ensemble techniques (SRP, LB, and PWPAE). The throughput of the proposed MSANA model is also at a high level (283.29 samples per second).

\begin{figure}[!t]
\centerline{
\includegraphics[width=7.0cm]{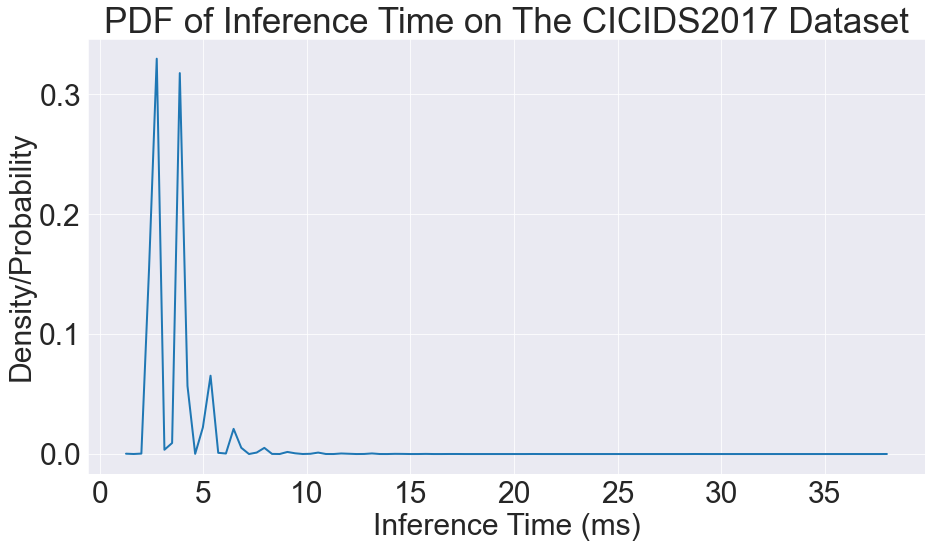}}
\caption{Probability Density Function (PDF) of the inference time of the proposed framework on the CICIDS2017 dataset.}
\label{cicpdf}
\end{figure}

\begin{figure}[!t]
\centerline{
\includegraphics[width=7.0cm]{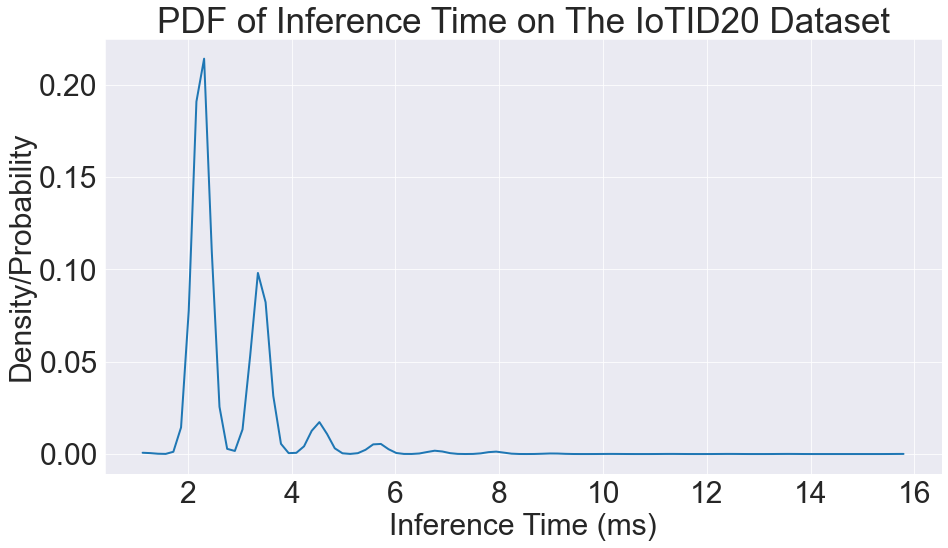}}
\caption{Probability Density Function (PDF) of the inference time of the proposed framework on the IoTID20 dataset.}
\label{iotidpdf}
\end{figure}

\begin{table}[]
\centering%
\caption{The Contributors to The Inference Time in The Proposed MSANA Framework}
\setlength\extrarowheight{1pt}
\scalebox{0.85}{
\begin{tabular}{|>{\centering\arraybackslash}m{7em}|>{\centering\arraybackslash}m{8.3em}|>{\centering\arraybackslash}m{7.8em}|>{\centering\arraybackslash}m{6.6em}|}
\hline

\textbf{Category}                            & \textbf{Procedure/ Component}            & \textbf{Inference Time /Latency on CICIDS2017 (ms)} & \textbf{Inference Time /Latency on IoTID20 (ms)}  \\ 
\hline
\multirow{2}{7em}{\centering Dynamic Data Pre-Processing} & Data Balancing                    & 0.09                                               & 0.08                                             \\ 
\cline{2-4}
                                             & Data Normalization                & 0.02                                               & 0.02                                             \\ 
\hline
\multirow{2}{7em}{\centering Dynamic Feature Selection}   & Concept Drift Detection           & 0.002                                              & 0.002                                            \\ 
\cline{2-4}
                                             & Feature Selection                 & 0.05                                               & 0.04                                             \\ 
\hline
\multirow{5}{7em}{\centering Model Learning}              & Online Base Model Learning        & 3.21                                               & 2.54                                             \\ 
\cline{2-4}
                                             & Dynamic Model Selection           & 0.01                                               & 0.01                                             \\ 
\cline{2-4}
                                             & Online Ensemble Model Development & 0.15                                               & 0.12                                             \\ 
\hline
Overall                                      & All                               & 3.53                                               & 2.81                                             \\
\hline

\end{tabular}
}
\label{contributor}
\end{table}

The evaluation results for the online learning models on the IoTID dataset are shown in Fig. \ref{iotid} and Table \ref{results}. Similarly, the two leader models, ARF-ADWIN and ARF-EDDM, achieve high F1-scores of 99.15\% and 99.13\%. For the follower models, EFDT and KNN-ADWIN are selected to build the initial ensemble model due to their better performance when compared with SAM-KNN and OPA. The proposed MSANA method achieves the second-highest accuracy of 98.88\% and the second-highest F1-score of 99.41\%. Although the proposed MSANA method has a slightly lower accuracy than the existing PWPAE method (98.88\% versus 98.90\%), it has much shorter inference time (2.81 ms versus 9.08 ms) and much higher throughput (355.87 versus 110.13 samples per second) than PWPAE. This is because, in comparison to the PWPAE method, the proposed MSANA method uses a window-based strategy and selects lightweight base learners with greater computational speeds to build the ensemble model. Thus, the proposed MSANA method is still the best model in terms of balancing model performance and efficiency among the evaluated models.

Moreover, to evaluate the efficiency and the feasibility of the proposed framework, the Probability Density Functions (PDF) of the inference time/latency of the proposed framework on the CICIDS2017 and IoTID20 datasets are shown in Figs. 4 and 5, respectively. Figures 4 and 5 illustrate that the vast majority of data samples were processed within 6 ms and 4 ms, respectively. There are several data packets in the two datasets that took a relatively long time (up to 16 - 38 ms) to be processed, as the learning models were updated after detecting concept drifts. Due to the low frequency of the occurrence of concept drift in real-world IoT data streams, only a tiny percentage of real-world IoT data would take a relatively long time to process. Additionally, even if the concept drift frequency of the experiments is high, the average inference time/latency of the proposed framework is still at a low level (3.53 ms and 2.81 ms on the two datasets), as shown in Table I. Please note that this inference time will be reduced dramatically when using high-performance computing devices that are usually deployed on IIoT cloud servers.

Lastly, to identify the contributors to the inference time/latency, the inference time of each component of the proposed MSANA framework is computed and provided in Table II. As shown in Table II, the inference time of most components/procedures of the proposed framework is shorter than 0.1 ms, and the proposed W-PWPAE ensemble model has low latencies on the two datasets (0.15 ms and 0.12 ms) due to its low computational complexity. The primary contributor to the inference time is the online base model learning procedure, since it takes time for multiple online learners to learn the data streams. Thus, the proposed framework itself has high efficiency, which can be further improved by replacing the base learners with lightweight learning models with lower complexity.

\section{Model Deployment and Expected Performance}
\begin{figure}[!t]
\centerline{
\includegraphics[width=\columnwidth]{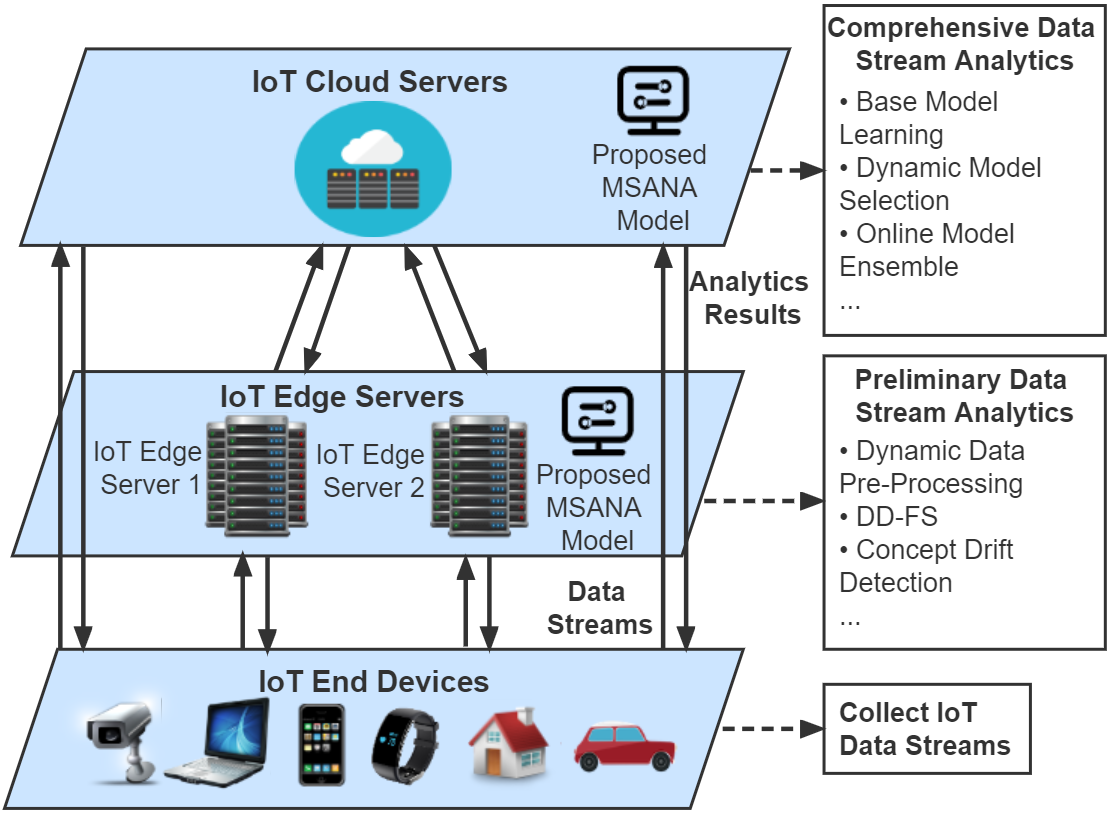}}
\caption{The potential deployment of the proposed model in IoT systems.}
\label{iotfig}
\end{figure}

In this section, the potential deployment and the expected performance of the proposed MSANA framework in practical IIoT systems are discussed. Typical IoT system components related to IoT data analytics include IoT end devices for IoT data stream collection, IoT edge servers for preliminary and local data analytics, and IoT cloud servers for comprehensive data analytics \cite{iotyu}, as shown in Fig. 6. 

The proposed MSANA model can be deployed on both IoT cloud and edge servers for IoT data stream analytics, also known as a collaborative computing architecture \cite{colla}. Firstly, IoT data streams can be analyzed locally on edge servers or end devices via edge computing. IoT edge devices enable fast data processing by avoiding long-distance data transmission and reducing latency, but they usually have limited computational resources due to their low-cost requirements. To deploy the proposed MSANA model, edge servers can provide preliminary and fundamental IoT data stream analytics tasks locally, such as dynamic data pre-processing and feature selection, as they can improve data quality and do not require many computational resources. 

On the other hand, the proposed MSANA model can be deployed on IoT cloud servers for complex IoT data stream analytics. IoT cloud servers usually have multiple cloud machines with strong computational power and resources, so they can perform comprehensive data analytics tasks by cloud computing \cite{iotyu}. As analytics model learning tasks are time-consuming and resource-consuming tasks, these procedures inside the proposed MSANA framework can be implemented on IoT cloud servers, including base model learning, dynamic model selection, and online model ensemble. Thus, deploying the proposed framework on both IoT cloud and edge servers using collaborative computing can provide IoT users and devices with reliable and fast services. 

The estimated performance of the proposed MSANA framework in practical IIoT systems is discussed from both the learning effectiveness and efficiency perspectives. Firstly, from the prediction accuracy/effectiveness perspective, the proposed MSANA framework could achieve the same or similar level of prediction performance (e.g., accuracy and F1-scores) as our experimental results, as real-world IoT cloud servers are usually high-end machines whose computing power is higher than the power of the computer used in our experiments. Although a small percentage of network packets might be lost due to network transmissions, our proposed framework could achieve high accuracy and F1-scores even with very few data samples, as shown in Figs. \ref{cic} and \ref{iotid}. This is primarily due to the use of data sampling and online learning methods that do not require a large volume of data. 

Secondly, from the efficiency perspective, the response time or latency of data stream analytics might increase due to data transmission delay or network congestion. Nevertheless, due to the high speed of current 4G and 5G networks (average speed $>$ 30 megabits per second (Mbps)), the transmission delay of each data packet would be much lower than 0.1 ms \cite{5g}. Since the average inference time of the proposed MSANA framework on the two benchmark datasets is less than 3.6 ms, the expected total latency would be lower than 3.7 ms. Moreover, to evaluate the drift adaptability of the proposed framework, several concept drifts occurred in a short time period (less than one hour) in the experiments, while concept drifts would occur less often in practical settings. Since the only time-consuming procedure of the proposed framework, model updating/adaptation, only need to be conducted when concept drift occurs, the estimated latency in practical applications would be lower than the latency in our experiments (3.53 ms). Moreover, the latency will be reduced dramatically when using high-performance computing devices that are usually deployed on cloud servers. As the latency requirement of real-world IIoT systems is 10 to 100 ms, the proposed framework could still achieve real-time analytics \cite{iotyu}.

\section{Conclusion}
Network automation technologies have drawn much attention to the development of IIoT applications in Industry 4.0 and 5.0. AI and ML algorithms are critical techniques for the automation of network data analytics, which is a key component of IIoT network automation. However, owing to the dynamic nature of IIoT environments, IIoT data is usually large data streams that are continuously generated and changed. As a consequence, concept drift issues often occur as a result of IIoT data distribution changes, causing learning model deterioration. In this paper, we proposed a comprehensive automated data analytics framework that is capable of processing continuously evolving IIoT data streams while dynamically adapting to concept drifts. The proposed framework consists of dynamic data pre-processing, drift-based dynamic feature selection, dynamic model selection, and online model ensemble using a novel W-PWPAE approach. According to the model performance evaluations on two benchmark IIoT streaming datasets, IoTID20 and CICIDS2017, the proposed framework is capable of effectively processing dynamic IIoT streams with higher accuracy of 98.88\% and 99.32\%, respectively, than other state-of-the-art methods. In future work, the proposed framework will be deployed and evaluated in real-world IIoT networks using more QoS parameters. Additionally, the proposed framework can be extended by introducing more advanced data learning techniques to further improve learning performance and speed.

\balance

\begin{IEEEbiography}[{\includegraphics[width=1in,height=1.25in,clip,keepaspectratio]{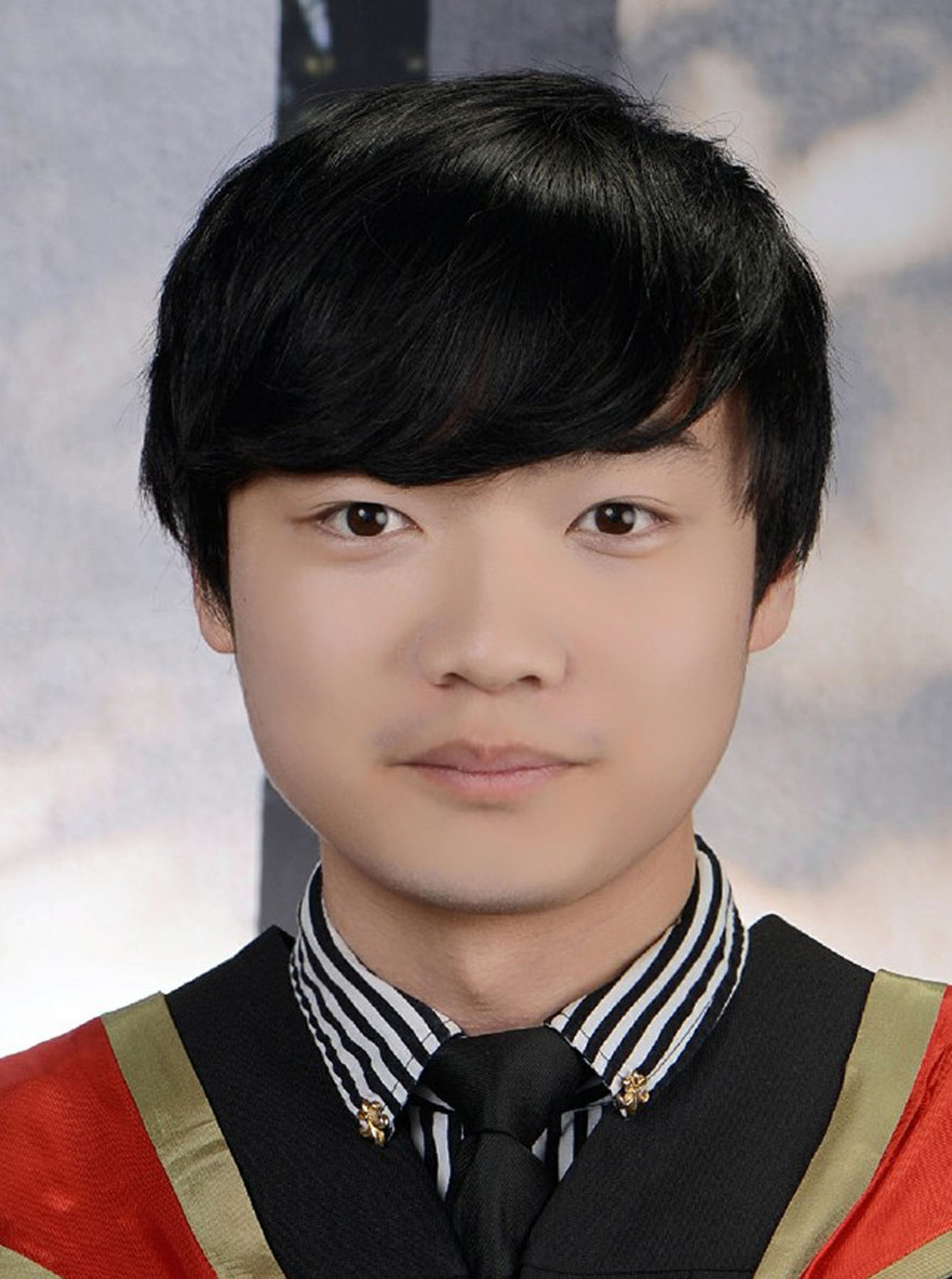}}]{Li Yang}received his Ph.D. in Electrical and Computer Engineering from Western University, London, Canada, in 2022, and his MASc. degree in Engineering from the University of Guelph, Guelph, Canada, in 2018. Currently, he is a Postdoctoral Associate and Sessional Lecturer in the OC2 Lab at Western University.  His research interests include cybersecurity, machine learning, AutoML, deep learning, IoT data analytics, anomaly detection, and time series analytics.
\end{IEEEbiography}

\begin{IEEEbiography}[{\includegraphics[width=1in,height=1.25in,clip,keepaspectratio]{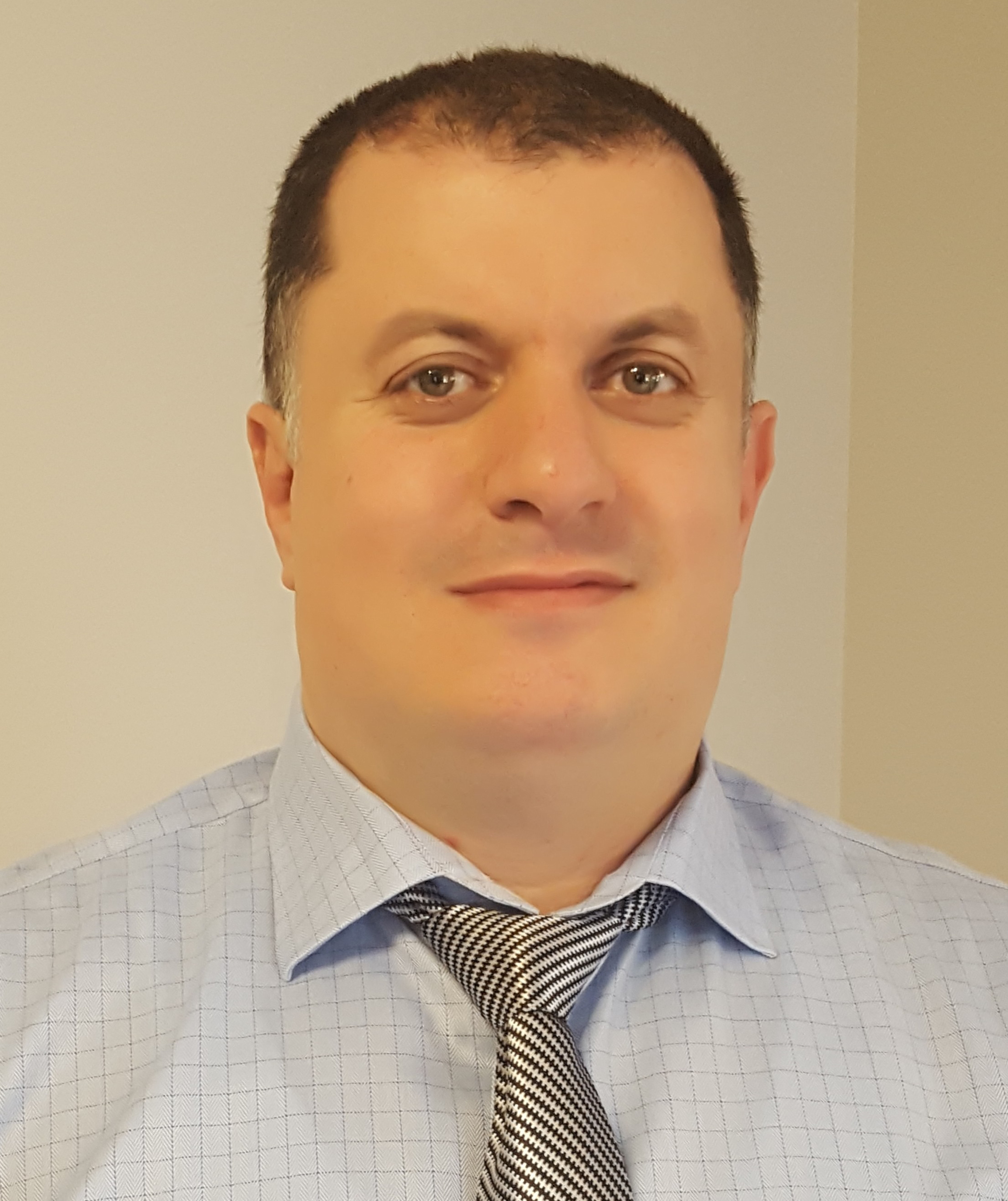}}] {Abdallah Shami}is currently a Professor in the Electrical and Computer Engineering Department and the Acting Associate Dean (Research) of the Faculty of Engineering, Western University, London, ON, Canada, where he is also the Director of the Optimized Computing and Communications Laboratory. Dr. Shami has chaired key symposia for the IEEE GLOBECOM, IEEE International Conference on Communications, and IEEE International Conference on Computing, Networking and Communications. He was the elected Chair for the IEEE Communications Society Technical Committee on Communications Software from 2016 to 2017 and the IEEE London Ontario Section Chair from 2016 to 2018. He is currently an Associate Editor of the IEEE Transactions on Mobile Computing, IEEE Internet of Things, and IEEE Communications Surveys and Tutorials journals.
\end{IEEEbiography}

\end{document}